\documentclass[conference]{IEEEtran}
\IEEEoverridecommandlockouts
\usepackage{cite}
\usepackage{amsmath,amssymb,amsfonts}
\usepackage{algorithmic}
\usepackage{graphicx}
\usepackage{textcomp}
\usepackage{xcolor}
\usepackage{algorithm}
\usepackage{latexsym}
\usepackage{amssymb}
\usepackage{amsmath}
\usepackage{amsthm}
\usepackage{booktabs}
\usepackage{enumitem}
\usepackage{color}
\usepackage{subfigure}
\usepackage{mdwlist}
\usepackage{multicol, multirow}
\usepackage{bbding}
\usepackage{pifont}
\usepackage{utfsym}
\usepackage{cellspace}
\usepackage{algorithm}
\usepackage{algorithmic}

\usepackage{algorithmic}
\usepackage{diagbox}
\usepackage{xspace}
\usepackage{url}

\newcommand{\mypara}[1]{{\noindent\textbf{#1}}}

\newcommand\ourmethod{StrAD\xspace}
\def\BibTeX{{\rm B\kern-.05em{\sc i\kern-.025em b}\kern-.08em
    T\kern-.1667em\lower.7ex\hbox{E}\kern-.125emX}}
\begin{document}

\title{An Improved Time Series Anomaly Detection by Applying Structural Similarity}


\author{
\IEEEauthorblockN{Tiejun Wang\IEEEauthorrefmark{1},
Rui Wang\IEEEauthorrefmark{1},
Xudong Mou\IEEEauthorrefmark{1},
Mengyuan Ma\IEEEauthorrefmark{2},
Tianyu Wo\IEEEauthorrefmark{2},
Renyu Yang\IEEEauthorrefmark{2},
Xudong Liu\IEEEauthorrefmark{1}}

\IEEEauthorblockA{\IEEEauthorrefmark{1}School of Computer Science and Engineering, Beihang University, Beijing, China } 
\IEEEauthorblockA{\IEEEauthorrefmark{2}School of Software, Beihang University, Beijing, China}
\IEEEauthorblockA{ \{wtj, ruiking, mxd, mamy, woty, renyuyang, liuxd\}@buaa.edu.cn\} }\\
\thanks{Corresponding authors: Tianyu Wo (woty@buaa.edu.cn), Renyu Yang (renyuyang@buaa.edu.cn).}
}
\maketitle

\begin{abstract}
Effective anomaly detection in time series is pivotal for modern industrial applications and financial systems. Due to the scarcity of anomaly labels and the high cost of manual labeling, reconstruction-based unsupervised approaches have garnered considerable attention.
However, accurate anomaly detection remains an unsettled challenge, since the optimization objectives of reconstruction-based methods merely rely on point-by-point distance measures, ignoring the potential structural characteristics of time series and thus failing to tackle complex pattern-wise anomalies. 
In this paper, we propose \ourmethod, a novel structure-enhanced anomaly detection approach to enrich the optimization objective by incorporating structural information hidden in the time series and steering the data reconstruction procedure to better capture such structural features.
\ourmethod accommodates the trend, seasonality, and shape in the optimization objective of the reconstruction model to learn latent structural characteristics and capture the intrinsic pattern variation of time series. The proposed structure-aware optimization objective mechanism can assure the alignment between the original data and the reconstructed data in terms of structural features, thereby keeping consistency in global fluctuation and local characteristics.
The mechanism is pluggable and applicable to any reconstruction-based methods, enhancing the model sensitivity to both point-wise anomalies and pattern-wise anomalies. Experimental results show that \ourmethod improves the performance of state-of-the-art reconstruction-based models across five real-world anomaly detection datasets.
\end{abstract}
\begin{IEEEkeywords}
Time series, Anomaly detection, Structural Similarity
\end{IEEEkeywords}

\section{Introduction}\label{sec:intro}
Artificial intelligence (AI), Deep Learning (DL), and multimedia technology have driven advancements across all walks of life, propelling multimedia wearable devices towards greater intelligence. Leveraging their mobility, these smart wearables continuously record a vast amount of users' time-series life data -- such as location, sleep patterns, and heart rate -- during use. This capability makes them the preferred choice for individuals focused on health management. Consequently, detecting anomalies in this time-series data has become a major research focus.
However, anomalies are intrinsically sparse and obtaining accurately labeled data is intractable, making it only resolvable by unsupervised learning tasks~\cite{schmidl2022anomaly}.

\setlength{\textfloatsep}{2mm} 
\begin{figure}[t]
    \centering
    \includegraphics[width=0.85\linewidth]{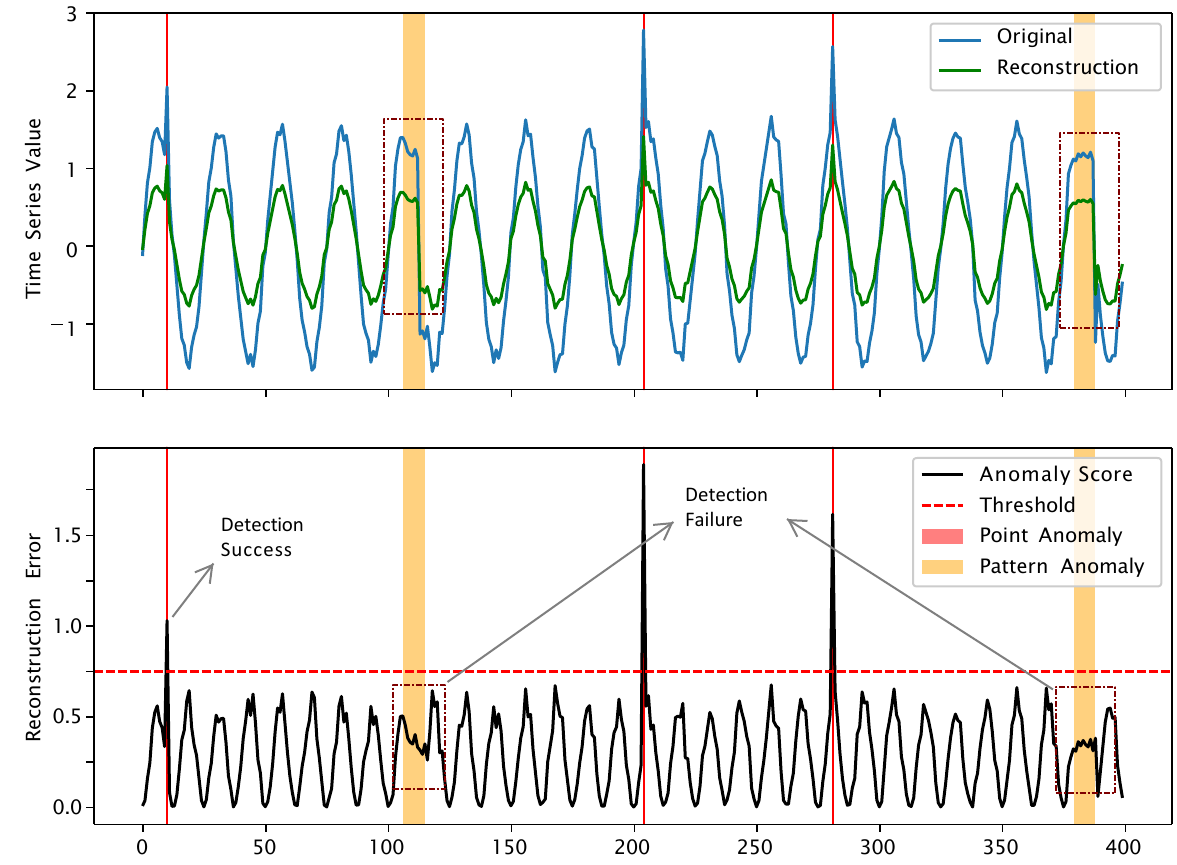}
    \caption{Reconstruction results and Reconstruction errors of original sequences containing point- and pattern-wise anomaly from dataset NeurIPS-TS  \cite{lai2021revisiting}, with Mean Squared Error as the optimization objective.}
    \label{1.1}
\end{figure}

Unsupervised DL models learn intrinsic features through reconstruction and prediction, with the normality assumptions of capturing normal patterns \cite{zamanzadeh2024deep,pang2021deep}.
Reconstruction-based methods such as  Anomaly Transformer \cite{xu2022anomaly}, AOC \cite{mou2023deep}, SensitiveHUE \cite{feng2024sensitivehue}, and LSTM-ED \cite{malhotra2016lstm} encode normal data in the latent space and minimize the residual between the decoded data after reconstruction and the actual data. They typically assume a low reconstruction probability of anomalies~\cite{wang2023deep}, due to their rare and unpredictable nature.

Noticeably, time series usually exhibit a variety of morphological patterns, such as wave-like, step-wise, or curved forms, and embedding-wise rich structural information including trends and periodicity \cite{lai2021revisiting}. 
Reconstruction-based methods typically learn to capture the underlying temporal dynamics by optimizing an objective function that solely measures point-by-point distances -- such as Mean Squared Error (MSE), Dynamic Time Warping (DTW), and Mean Absolute Error (MAE) -- yet entirely overlooking such structural features as trends and seasonal information, which in turn impeds accurate anormaly identification. Fig. \ref{1.1} shows the insufficient capability of handling pattern-wise anomalies, if only considering point-by-point distance optimization in the state-of-the-art reconstruction methods.


In this paper, we present \ourmethod{}, a novel \underline{Str}ucture-enhanced \underline{A}nomaly \underline{D}etection model that enriches the optimization objective by characterizing structural information of time series and guiding the reconstruction model to capture such structural features. The core idea is to redesign the loss function to measure the structural similarity between the original and reconstructed data. \ourmethod{} breaks down the calculation of similarity from three distinct yet interconnecting aspects -- trend, seasonality, and shape. 
To prevent the effects of real-world time series noises, \ourmethod{} extracts trend and seasonality characteristics through Legendre Polynomials projection and Fast Fourier Transform, respectively.
As a local structural characteristic, the shape information is well elaborated in the loss to avoid the excessive influence of high error points.
Inspired by structural modeling of time series, \ourmethod{} integrates the three structural components in a weighted manner to form a unified loss function.
\ourmethod{} not only guides models to learn structural features but also captures long-term global patterns and local variations of time series. 
Additionally, the loss calculation can be seamlessly integrated with conventional point-by-point loss functions, thereby enhancing both the robustness and the accuracy of anomaly detection in time series data. 
We conduct extensive experiments on five benchmark datasets, and experimental results show that \ourmethod{} consistently improves the performance of reconstruction-based methods in TSAD tasks. 
In particular, this work makes the following contributions:

\begin{itemize*}
    \item Introducing structural similarity, for the first time, as a new driving force that navigates the optimization objective for reconstruction models, without any additional architectural modifications.
    \item Redefining a structure-aware loss function that takes the full advantage of trend, seasonality, and shape to measure the difference between the origin and reconstruction data. 
    \item Conducting extensive experiments on five real-world datasets, both univariate and multivariate, and three state-of-the-art architectures to show the efficacy of \ourmethod{}.
\end{itemize*}
\section{Related Work} 
\label{sec:related}

\mypara{Time Series Anomaly Detection (TSAD)}. TSAD is important for various research fields and applications. Deep learning based approaches can effectively learn expressive representations of complex data and are competitive in tackling challenging anomaly detection problems \cite{wang2024deep}.
LSTM-ED \cite{malhotra2016lstm} proposes an encoder-decoder scheme based on long-short-term memory networks that learns to reconstruct normal time-series behavior.
COCA \cite{wang2023deep} and AOC \cite{mou2023deep} integrate contrastive learning and autoencoder with one-class classification by constructing a minimal hypersphere to distinguish anomalies. RoCA \cite{mou2025roca} introduces a multi-hypothesis framework to enhance anomaly detection. TranAD \cite{tuli2022tranad} uses attention-based sequence encoders to swiftly perform inference with the knowledge of the broader temporal trends in the data. 
Based on the diffusion model, D${^3}$R \cite{wang2023drift} utilizes data-time mix-attention to dynamically decompose long-period multivariate time series, to overcome the limitation of the local sliding window. Furthermore, foundation models such as UniTS \cite{gao2024units}, ALFA \cite{zhu2024llms}, and One Fits All \cite{zhou2023one} introduce new model frameworks to TSAD tasks. These foundation models are trained on large-scale, diverse datasets to perform data analysis tasks by capturing broad and universal features, which exhibit remarkable zero-shot capabilities when analyzing new data domains in prediction and classification tasks.
However, most of them improve detection performance through model architectural improvements without a special focus on optimizing TSAD-specific objectives.

\mypara{Reconstruction-based Anomaly Detection}.  Most of the existing studies formulate TSAD tasks as an unsupervised problem that can be resolved by reconstruction-based approaches. One direction is improving network architecture~\cite{wang2023deep, wang2024cutaddpaste}.
AMDN \cite{xu2015learning} combines both the benefits of traditional early fusion and late fusion strategies to exploit the complementary information of both appearance and motion patterns. UODA \cite{lu2017unsupervised} uses auto-encoder models to capture the intrinsic difference between outliers and normal instances. OmniAnomaly \cite{su2019robust} proposes a stochastic recurrent neural network for multivariate TSAD that works robustly for various devices. 
Another research direction lies in improving the anomaly score instead of using the original reconstruction error. FreDF \cite{wang2024fredf} and PS \cite{kudrat2025patch} overcome the limitations of traditional optimization objectives by enhancing frequency domain information and patch localized statistical properties. NPSR \cite{lai2023nominality} introduces a nominality score calculated from the ratio of the combined value of the reconstruction errors.
Anomaly Transformer \cite{xu2022anomaly} introduces a new Anomaly-Attention mechanism to compute the association discrepancy.
To enhance the sensitivity of normal patterns, SensitiveHUE \cite{feng2024sensitivehue} proposes a probabilistic network by implementing both reconstruction and heteroscedastic uncertainty estimation.
While they can somewhat improve the detection performance, such works largely overlook the structural characteristics inherent in time series.
Our work aims to propose a structure-aware optimization objective for time series anomaly detection, which learns latent structural features to capture time series pattern variation.

\begin{figure*}[!ht] 
    \centering
    \subfigure{\includegraphics[width=0.85\textwidth]{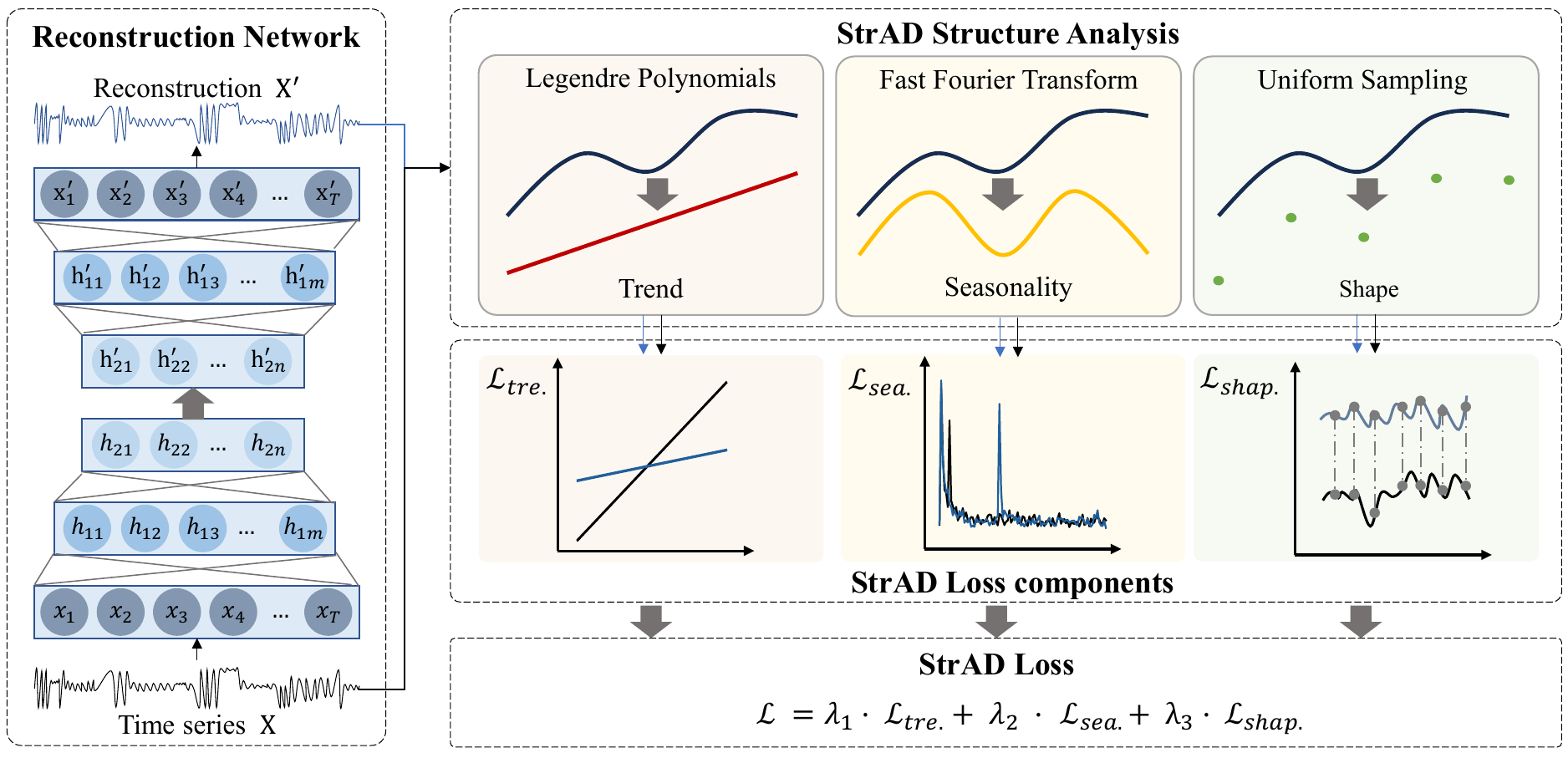}}
    \caption{Overview of the structure-aware \ourmethod{} optimization objective, which measures structure similarity between origin and reconstructed time series data by combining trend, seasonality, and shape. }
    \label{3.1}
\end{figure*}
\section{Methodology} 

\subsection{Problem Definition}
Given a time series $\mathcal S$ = \{$x_1$, $x_2$,..., $x_M$\} with length $\mathcal M$, where $x_i \in \mathbb{R}^d $ is a $d-$dimensional vector collected at timestamp $i$.
Conventional approaches typically use sliding windows with length $t$ to process $\mathcal S$ into subsequence set  $\mathcal D$ =  $\mathrm{ \{ X_1,X_2,...,X_N \}} $, where sample $ \mathrm{X_i =\{ x_1,x_2,...,x_t }$ \}is a subsequence of $\mathcal S$ with length $t$, and $ N$ is the number of the subsequence. We follow the structural modeling \cite{lai2021revisiting} to better represent the characteristics of time series.
\begin{equation}
   X = \Gamma(2\pi \omega T) + \Theta(T),
\label{eq3.1}
\end{equation}
where $T=\{1,2,...,t\}$.
$\Gamma(\cdot)$ and $\Theta(\cdot)$ denote the base shapelet and trend function describing the direction of $\mathrm{X_i}$.

Typically, a TSAD detection model calculates an anomaly score $\mathrm{S_i}$ for each $\mathrm{X_i}$. The higher $\mathrm{S_i}$ is, the more likely $\mathrm{X_i}$ is determined as anomalous. Reconstruction-based models learn the latent features of the input data ${\bf X}$ and compare the reconstructed data ${\bf X}^{\prime}$ with ${\bf X}$ to obtain the anomaly score. Such models are designed to fit the underlying parameter defined in Eq. \ref{eq3.1} to obtain the errors between the  ${\bf X}$ and ${\bf X}^{\prime}$. 
However, these models solely rely on point-by-point distance as their optimization objective and fall short in accurately fitting Eq. \ref{eq3.1} when the time series exhibits variations in trend or periodicity. Therefore, it is imperative to take into account such structural information as trend, periodicity, and shape in the calculation of the loss function. 

\subsection{Structural Characteristics}
At the core of \ourmethod{} is to analyze the structural characteristics of time series and introduce a structural similarity loss function to navigate the pattern learning of given time series from a structural perspective. As observed in Eq. \ref{eq3.1}, we break down the structural elements into trend, seasonality, and shape.
The shapelet function $\Gamma(\cdot)$ reveals that $\omega$ reflects the periodic nature of the time series and $\Gamma(\cdot)$ comprises both global and local characteristics. As illustrated in Fig. \ref{3.1}, we formulate a structure-aware loss function to accommodate  \textit{trend, seasonality, and shape}—to quantify the discrepancy in structural similarity between the original time series ${\bf X}$ and its reconstruction ${\bf X}^{\prime}$. 

\subsection{Structure Similarity Aware Loss Enhancement} 

We enrich the conventional loss calculation in a straight-forward yet working manner as below. 

\mypara{Trend.}
As illustrated in Eq. \ref{eq3.1}, the trend is abstracted as the function $\Theta(T)$ solely dependent on time $T$.
Continuous variation of a time series is typically manifested as rising or falling trend features, and such trend fluctuation gets even more significant as the time window increases.
To effectively capture the global trend variation, Legendre polynomials $\mathcal{P(\cdot)}$ are often employed owing to their capability of representing smooth variations over fixed intervals.
Accordingly, the trend term is calculated by projecting both the original and the reconstructed sequences onto the linear trend, and then by performing comparative calculations:

\begin{equation}
\begin{aligned}
\mathcal{L}_{trend} &= \mathcal{P}_n(\mathbf{X}) - \mathcal{P}_n(\mathbf{X}^{\prime}) \\
                    &= \sum_{k=0}^{n} a_k \cdot t^k - \sum_{k=0}^{n} b_k \cdot t^k ,
\end{aligned}
\label{eq02}
\end{equation}
where $a_k$ and $b_k$ represent the coefficients of the fitted polynomials of $\bf {X} $ and $\bf{X}^{\prime}$, respectively.
We advocate the use of $ n = 1$ in \ourmethod{}, which allows the trend term to represent potential long-term changes in the time series. 
Moreover, the trend fluctuations of time series typically evolve slowly and smoothly, while certain anomalous segments may cause abrupt deviations. 
To ensure training stability and improve error sensitivity, the trend item is processed, and then the trend loss is defined as:

\begin{equation}
\begin{aligned}
\mathcal{L}_{tre.} &= -\log \left( \left| \mathcal{P}_1(\mathbf{X}) - \mathcal{P}_1(\mathbf{X}^{\prime})  \right|_1 + \varepsilon \right)\\
                   &= -\log \left( \left|  a \cdot t -  b \cdot t \right|_1 + \varepsilon \right) ,
\end{aligned}
\label{eq03}
\end{equation}
where \( \varepsilon  \) is a small constant, $ \left| \cdot \right|_1$ denotes the element-wise $\ell_1$ norm, summing the absolute values of all elements.

\mypara{Seasonality.}
According to the shapelet function $\Gamma(\cdot)$ in Eq. \ref{eq3.1}, there are patterns of normal variation in time series that repeat at regular intervals, which have a potentially fixed frequency of variation, that is, seasonal characteristics $\omega$. 

However, real-world time series inevitably contain noise, making it intricate but pivotal to identify seasonal variations accurately. To tackle noisy time series, the Fast Fourier Transform (FFT) has been demonstrated effective in identifying underlying periodic patterns while effectively suppressing local disturbances caused by noise, owing to mapping the sequence from the time domain to the frequency domain.
We therefore apply FFT to capture the seasonality characteristics:
\begin{equation}
\mathcal{L}_{sea.} = \left| \mathcal{F}({\bf X}) - \mathcal{F}({\bf X}^{\prime}) \right|_1,
\label{eq04}
\end{equation}
where \( \mathcal{F}(\cdot) \) denotes the FFT algorithm. The seasonality feature represents the regular fluctuation of a time series. As one of the structure similarity measures, minimizing the difference in seasonality is helpful to align long-term variation. 

\mypara{Shape.}
Time series inherently contain structural shape characteristics, including peaks and valleys, which reflect the local fluctuation. Such features are usually composed of several data points. 
Therefore, the shape fluctuation information can be obtained by comparing the differences point by point across the time series. To diminish the amplification of local errors, it is helpful to apply the absolute value operation on these differences instead of squaring them, i.e., $\ell_1$ loss.
Doing so can mitigate overreaction to local deviations and prevent data points with excessive errors from dominating the optimization process. Accordingly, the final shape loss $\mathcal{L}_{shap.}$ is adjusted as follows:

\begin{equation}
\mathcal{L}_{shap.} = \sum_{t=1}^{T} \left| {x_t} - {x_t^{\prime}} \right|_1 .
\label{eq05}
\end{equation}

\mypara{Putting Them Together.} By aligning the similarity of trends and seasonality, the original data $\textbf{X}$ and reconstructed data $\textbf{X}^{\prime}$ maintain consistency in their global long-term variation patterns. 
Meanwhile, minimizing the shape discrepancies can make the alignment as much as possible between the two time series at the local detailed level.
Therefore, \ourmethod{} can effectively capture the intrinsic structure of time series by integrating all these elements above. In practice, by following the form of Eq. \ref{eq3.1}, the overall structure-aware loss function \ourmethod{} can be defined as a weighted combination of three components: 
\begin{equation}
\mathcal{L} = \lambda_{1} \cdot \mathcal{L}_{tre.} + \lambda_{2} \cdot \mathcal{L}_{sea.} + \lambda_{3} \cdot \mathcal{L}_{shap.} .
\label{eq06}
\end{equation}
$\lambda_{1}$, $\lambda_{2}$, and $\lambda_{3}$ are weight parameters controlling the relative contribution of each structure characteristic.
They can be dynamically updated according to the gradient magnitudes or manually adjusted based on domain-specific prior knowledge, according to the characteristics of the datasets.
Notably, the values of parameters $\lambda_{1}$, $\lambda_{2}$, and $\lambda_{3}$ should be to ensure that the relative contributions of different components remain within an appropriate range.
In the next section, we perform a sensitivity analysis to assess its influence on model performance.
By learning the global and local structural changes of time series, \ourmethod{} improves the sensitivity to point- and pattern-wise anomalies. The structure-awareness in the loss function is highly flexible and can be generalized in many other reconstruction-based models, without any additional modification.

\begin{algorithm}[tb]
\caption{Structure Similarity Guided Time Series Anomaly Detection}
\label{alg:algorithm}
\textbf{Input}: Time series $\bf{X}$ \\
\textbf{Parameter}: Encoder $E$, Decoder $D$\\
\textbf{Output}: Trained model $\mathcal{M}$ 

\begin{algorithmic}[1] 
\STATE Initialize weights of $E$, $D$
\FOR{each training iteration}
    \STATE Time series sample $\bf{X}$
    \STATE Encode latent representation: $\bf{H}$ $\leftarrow$ ${E(\bf{X})}$
    \STATE Reconstruct: $\bf{X}^{\prime}$ $\leftarrow$ $D(\bf{H})$
    \STATE Analyze structural characteristics:
    \STATE $\mathcal{L}_{tre.} = -\log \left( \left|  a \cdot t -  b \cdot t \right|_1 + \varepsilon \right)$
    \STATE $\mathcal{L}_{sea.} = \left| \mathcal{F}({\bf X}) - \mathcal{F}({\bf X}^{\prime}) \right|_1$
    \STATE $\mathcal{L}_{shap.} = \sum_{t=1}^{T} \left| {x_t} - {x_t^{\prime}} \right|_1 $
    \STATE Aggregate \ourmethod{} loss: 
    \STATE $\mathcal{L} = \lambda_{1} \cdot \mathcal{L}_{tre.} + \lambda_{2} \cdot \mathcal{L}_{sea.} + \lambda_{3} \cdot \mathcal{L}_{shap.}$
    \STATE Update $E, D$ to minimize $\mathcal{L}$
\ENDFOR
\STATE \textbf{return} Trained model $\mathcal{M}$
\end{algorithmic}
\end{algorithm}

\setlength{\tabcolsep}{1mm}
\begin{table}[!htb]
    \centering
    \small
    \caption{Summary of TSAD datasets.}

    \begin{tabular}{cccc}
    \toprule
    Dataset & Entity / Dim & Train / Test & Anomaly Rate (\%) \\
    \midrule
    AIOps & 29 / 1 & 3003631 / 2918412  & $2.26$ \\
    UCR & 250 / 1 & 5298699 / 14047567  & $0.51$ \\
    SWaT & 1 / 51 & 475185 / 449904  & $5.90$ \\
    WADI & 1 / 127 & 784556 / 172788  & $1.04$ \\
    ESA & 1 / 6 & 12273601 / 12273599  & $0.76$ \\
    \bottomrule
    \end{tabular} 
\label{datasetdiv}
\end{table}

\section{Experiments}
\label{sec:expersetup}

\begin{table*}[!htb]
\centering
\small
\caption{The RPA F1-score(\%), and PA F1-score(\%) for MSE learning objectives, and our proposed \ourmethod{} with three SOTA TSAD models on AIOps, UCR, SWaT, WADI, and ESA datasets. The \textit{Avg}(\%) and \textit{A.I.R.} mean the average and the average improvement rate of \ourmethod{} compared to MSE. The terms highlighted in bold indicate the best results.}
\begin{tabular}{c|cc|cc|cc|cc|cc|cc}
\toprule
\multicolumn{1}{c|}{Model} & \multicolumn{4}{c|}{Anomaly Transformer ~\cite{xu2022anomaly}}  & \multicolumn{4}{c|}{AOC ~\cite{mou2023deep}} & \multicolumn{4}{c}{SensitiveHUE ~\cite{feng2024sensitivehue}} \\
\cmidrule(lr){2-13}
\multicolumn{1}{c|}{Loss Function} & \multicolumn{2}{c|}{+ MSE} & \multicolumn{2}{c|}{+ \ourmethod{}} & \multicolumn{2}{c|}{+ MSE} & \multicolumn{2}{c|}{+ \ourmethod{}} & \multicolumn{2}{c|}{+ MSE} & \multicolumn{2}{c}{\textbf{+ \ourmethod{}}} \\
\cmidrule(lr){2-3} \cmidrule(lr){4-5} \cmidrule(lr){6-7} \cmidrule(lr){8-9} \cmidrule(lr){10-11} \cmidrule(lr){12-13}
\multicolumn{1}{c|}{Metrics} & RPA F1 & PA F1 & RPA F1 & PA F1 & RPA F1 & PA F1 & RPA F1 & PA F1 & RPA F1 & PA F1 & RPA F1 & PA F1 \\
\midrule
\multirow{4}{*}{} 
   AIOps  & 0.49 & 34.75 & \textbf{1.91} & \textbf{58.55} & 8.06 & 39.53 & \textbf{32.06} & \textbf{61.88} & 4.56 & 32.37 & \textbf{28.31} & \textbf{59.47} \\
   UCR & 2.40 & 11.67 & \textbf{2.82} & \textbf{20.52} & 16.43 & 31.96 & \textbf{18.90} & \textbf{33.77} & 11.55 & 22.29 & \textbf{15.69} & \textbf{26.71} \\
   SWaT & \textbf{33.33} & 92.03 & 5.14 & \textbf{97.93} & 29.79 & 82.33 & \textbf{30.51} & \textbf{83.03} & 33.33 & 83.41 & \textbf{38.60} & \textbf{87.16} \\
  WADI  & 1.69 & 52.91 & \textbf{2.02} & \textbf{93.62} & 0.68 & 11.26 & \textbf{5.71} & \textbf{11.39} & 31.58 & 28.12 & \textbf{35.71} & \textbf{58.47} \\
  ESA  & \textbf{5.20} &\textbf{99.78} & 3.01 & 99.06 & 17.14 & 89.48 & \textbf{79.25} & \textbf{97.59} & 20.44 & 80.30 & \textbf{88.14} & \textbf{97.94} \\
\midrule
\multicolumn{1}{c|}{\textit{Avg}} 
       & \textbf{8.62} & 58.23 & 2.98 & \textbf{73.94} & 14.42 & 50.91 & \textbf{33.29} & \textbf{57.53} & 20.29 & 49.30 & \textbf{41.29} & \textbf{65.95} \\
\multicolumn{1}{c|}{\textit{A.I.R.}} 
       & - & - & \textbf{0.40} & \textbf{0.45} & - & - & \textbf{2.83} & \textbf{0.15} & - & - & \textbf{1.83} & \textbf{0.48} \\
\bottomrule
\end{tabular}

\label{other_method_loss_function-results}
\end{table*}

\subsection{Experimental Setup}
\textbf{Datasets.}
We conduct a comprehensive experimental evaluation on 5 datasets from diverse domains, overcoming the limitation of using a single dataset with domain-specific features. 
The data statistics are summarized in Table \ref{datasetdiv}, and the details are as follows:
\begin{itemize*}
    \item \textbf{AIOps} \cite{AIOpsChallenge} encompasses 29 univariate sub-datasets of well-maintained business cloud KPIs from prominent Internet companies. It includes two types of anomalies: point-wise and pattern-wise, of which there are more point-wise anomalies.
    \item \textbf{UCR} \cite{wu2021current}  contains 250 univariate sub-datasets from various fields. Each contains only one anomaly segment, mainly pattern-wise anomaly types.
    \item \textbf{SWaT} \cite{mathur2016swat} is collected from a scaled-down water treatment testbed with 51 sensors over 11 days. During the last 4 days, anomalies were injected using diverse attack methods. It contains several long-term pattern-wise anomalies.
    \item \textbf{WADI} \cite{ahmed2017wadi} is acquired from a reduced city water distribution system with 127 sensors and actuators, with 14 days collected under normal status and 2 days under attack scenarios. Unlike SWaT, it contains shorter, long-term pattern-wise anomaly segments.
    \item \textbf{ESA} \cite{kotowski2024european} includes annotated real-life telemetry of three different missions from the European Space Agency. As a large multivariate dataset, it contains several long-term pattern-wise anomalies.
\end{itemize*}

\noindent\textbf{Metrics.}
We validate our approach in terms of anomaly detection capability and detection performance improvement.

i) \textit{Anomaly Detection.}
Currently, there is no unified standard for the selection of evaluation metrics for time series anomaly detection performance.
To evaluate the fairness of each metric, \cite{wang2024cutaddpaste} conducts analysis and discussion for specific evaluation metrics.
Compared with other metrics, the RPA metrics treat the entire abnormal segment as a single sample, which is consistent with the number of anomalies.
Therefore, this paper chooses RPA Precision, Recall, and F1-score to evaluate the detection performance of all models.
Note that this paper reports on the metrics for the entire dataset, which is a weighted average of the RPA F1-score for each sub-dataset:
\begin{equation}
   \rm F1_{\rm entire} = \sum_{i=1}^{M} \frac{e_i}{E} F1_i,
\end{equation}
where $M$ is the number of sub-datasets, $E$ is the total number of anomaly segments for the entire dataset, and $e_i$ is the number of anomaly segments of the $i$-th sub-dataset.

ii) \textit{Performance Improvement.}
Based on the RPA F1, we evaluate the performance of models employing different optimization objectives.
Taking MSE as the benchmark, we calculate the average (\textit{Avg.Improved}) and the average improvement rate(\textit{A.I.R.}) of different learning objectives on all datasets as follows:

\begin{equation}
\footnotesize
  Avg.Improved = \frac{1}{Q} \times \sum_{i=1}^{Q} ( {\rm F1}_{i}^{\ast} - {\rm F1}_{i}^{\rm MSE} ) ,
\end{equation}

\begin{equation}
\footnotesize
  A.I.R. = \frac{1}{Q} \times \sum_{i=1}^{Q}  \frac{( {\rm F1}_{i}^{\ast} - {\rm F1}_{i}^{\rm MSE} ) }{{\rm F1}_{i}^{\rm MSE} },
\end{equation}
where $Q$ represents the number of datasets, $\ast$ represents other optimization objectives.
For a comprehensive comparative analysis, we also report the results of the PA F1-score.

\begin{table*}[!htb]
\centering
\small
\caption{Comparison between \ourmethod{} and other loss functions on AIOps, UCR, SWaT, WADI, and ESA datasets, based on the SensitiveHUE model. The terms highlighted in bold indicate the best results, while the second-best ones are underlined.}
\begin{tabular}{c|c|cc|ccc|c|c}
\toprule
\multicolumn{2}{c|}{\multirow{2}{*}{\diagbox{\textbf{Loss Function}}{\textbf{Dataset}}} }  & \multicolumn{2}{c|}{Univariate} &  \multicolumn{3}{c|}{Multivariate} & \multicolumn{1}{c|}{\multirow{2}{*}{\textit{Avg.Improved}}} & \multicolumn{1}{c}{\multirow{2}{*}{\textit{A.I.R.}}} \\
\cmidrule(lr){3-7}
 \multicolumn{2}{c|}{}  & AIOps & UCR & SWaT & WADI & ESA & \multicolumn{1}{c|}{} & \multicolumn{1}{c}{} \\
\midrule
\multirow{2}{*}{MSE} 
& RPA F1  & 4.56 & 11.55  & 33.33 & 31.58 & 20.44 & - & - \\
& PA F1  & 32.37 & 22.29 & 83.41 & 28.12 & 80.30 & - & - \\
\midrule
\multirow{2}{*}{MSE + \ourmethod{}} 
& RPA F1  & \underline{27.50} & 14.89 & \underline{36.92} & \underline{34.78} & 21.21 & \underline{6.77} & \underline{1.11} \\
& PA F1  & \underline{58.83} & 25.52 & \underline{86.60} & \textbf{60.9} & 80.54 & \underline{13.18} & \underline{0.43} \\
\midrule
\multirow{2}{*}{DTW} 
& RPA F1  & 5.52 & 15.14 & 29.79 & 11.11 & 18.62 & -4.26 & -0.06 \\
& PA F1  & 25.85 & \underline{26.94} & 82.33 & 32.11 & 85.14 & 1.18 & 0.04 \\
\midrule
\multirow{2}{*}{DTW + \ourmethod{}} 
& RPA F1  & 15.69 & \textbf{16.03} & 31.11 & 12.5 & \underline{25.00} & -0.23 & 0.48 \\
& PA F1  & 51.15 & \textbf{28.20} & 82.53 & 29.74 & \underline{93.98} & 7.82 & 0.21 \\
\midrule
\multirow{2}{*}{\ourmethod{}} 
& RPA F1  & \textbf{28.31} & \underline{15.69} & \textbf{38.60} & \textbf{35.71} & \textbf{88.14} & \textbf{21.00} & \textbf{1.83} \\
& PA F1  & \textbf{59.47} & 26.71 & \textbf{87.16} & \underline{58.47} & \textbf{97.94} & \textbf{16.65} & \textbf{0.48} \\
\bottomrule
\end{tabular}

\label{other_loss_function-results}
\end{table*}

\noindent \textbf{Baselines.}
We compare \ourmethod{} with other learning objectives: Mean Squared Error (MSE) and Dynamic Time Warping (DTW) \cite{sakoe1978dynamic}.
For the DTW learning objective, we adopt the DTW algorithm provided by \cite{lee2021soft_dtw_loss} that supports deep model training.
To comprehensively evaluate each loss function, we select three state-of-the-art (SOTA) reconstruction TSAD models, which include LSTM-based model: AOC \cite{mou2023deep}; Transformer-based model: Anomaly Transformer \cite{xu2022anomaly}, and SensitiveHUE \cite{feng2024sensitivehue}.

\mypara{Implementation}. For different learning objectives and other variants, we repeat each experiment with the same model and dataset.
We use the official implementations of each TSAD model from their respective repositories to obtain the best result by running 10 times.
Furthermore, for fair evaluation, only the optimization objective for reconstruction modules is modified in this experiment, while the network architecture and training strategy of each model remain unchanged.
All the models and loss functions are built with Pytorch and Merlion \cite{bhatnagar2021merlion}, and trained on an NVIDIA Tesla V100 GPU.

\subsection{Overall Performance}
Table \ref{other_method_loss_function-results} reports the entire performance of three SOTA TSAD models, by adopting \ourmethod{} or by default the MSE as loss function, respectively, on five real-world datasets.
Observably, for most cases, \ourmethod{} consistently outperforms the MSE loss function, only being slightly inferior to the MSE-based Anomaly Transformer on SWaT and ESA datasets.
Most notably, \ourmethod{}-based AOC improves average RPA F1 from 14.42\% to 33.29\% (18.87+\%), and \ourmethod{}-based SensitiveHUE improves RPA F1 from 20.29\% to 41.29\% (+21\%). Such improvement derives from \ourmethod{}'s ability to learn structural features within the time series, which can boost the model's sensitivity to anomalies.

To further evaluate the generality and flexibility of \ourmethod{} loss, we evaluate the \ourmethod{} against MSE, DTW, and integration with \ourmethod{} as a loss function.
The results are detailed in Table \ref{other_loss_function-results}.
We observe that the \ourmethod{} loss gets the best \textit{A.I.R.} 1.83 and 0.48 on five public datasets, respectively, in terms of RPA F1 and PA F1.
From the perspective of each dataset, compared with other loss functions, \ourmethod{} also achieves a remarkable improvement. 
In addition, MSE + \ourmethod{} and DTW + \ourmethod{} loss obtain the higher RPA F1 27.50\% and 16.03\% on AIOps and UCR dataset, respectively, which confirms that \ourmethod{} not only achieves superior performance when used as a standalone optimization objective, but also serves as a complementary component to enhance conventional objectives such as MSE or DTW.
This is due to its ability to capture structural features, thereby overcoming the limit of the point-by-point distance loss function.
These results highlight the potential of \ourmethod{} as a flexible plug-and-play strategy to enhance various reconstruction-based TSAD models.

Further analysis, as advocated by \cite{wang2024cutaddpaste,kim2022towards}, \ourmethod{} performance is not completely uniform on the two metrics of PRA and PA.
For instance, the RPA F1 obtained by Anomaly Transformer based on \ourmethod{} is not ideal on SWaT, but it performs well for PA F1. 
Through the analysis of the results presented in Table \ref{other_method_loss_function-results} and Table \ref{other_loss_function-results}, we argue that RPA provides a fairer evaluation and advocate the use of metric RPA instead of metric PA.


\subsection{Ablation Study}
To study the relative importance of each component of the \ourmethod{} method, we compare the performance for each variant of the \ourmethod{} optimization objective to observe how it affects the performance on AIOps and SWaT datasets.
The results are detailed in Table \ref{ablation}, where SensitiveHUE is employed as the anomaly detection model.
Overall, the \ourmethod{} optimization objective consistently outperforms the other six variant learning objectives, indicating both the effectiveness and importance of each component in our approach.
Upon further analysis, we observed that although \ourmethod{} demonstrated consistently strong overall performance,  certain variant optimization objectives also achieved remarkable results - for instance, the shape-based component on SWaT, as well as the trend-seasonality-based variants on AIOps.
This can be primarily attributed to the varying sensitivities of the six variants to different anomaly types, as well as the differences in the dominant anomaly types across the datasets.

Given that the SWaT multivariate dataset includes a substantial number of complex pattern-wise anomalies, the shape component effectively captures the local dynamic characteristics of time series, so that the shape-based variant achieves strong performance.
For the AIOps univariate datasets,  which include point- and pattern-wise anomalies, the trend and seasonality components are more sensitive to these two datasets. 
Although in the case of domain-specific datasets, concentrating on a single structural feature could be an effective strategy, the performance of relying on a single structural feature lacks generalizability.
For instance, although the shape component achieves outstanding results on the SWaT dataset, its performance on the AIOps datasets remains relatively modest.
In contrast, the \ourmethod{} optimization objective exhibits consistently impressive performance across the datasets.

\begin{table}[tb]
\centering
\small
\caption{ Ablation study results for each variant of the \ourmethod{} optimization objective. The terms highlighted in bold indicate the best results.}
\begin{tabular}{ccc|cc|cc} 
\toprule
\multicolumn{1}{c}{\multirow{2}{*}{Trend }} & \multicolumn{1}{c}{\multirow{2}{*}{Seasonality}} & \multicolumn{1}{c|}{\multirow{2}{*}{Shape}} & \multicolumn{2}{c|}{AIOps} &\multicolumn{2}{c}{SWaT} \\
\cline{4-7}
\multicolumn{1}{c}{} & \multicolumn{1}{c}{} & \multicolumn{1}{c|}{} & RPA F1 & PA F1  & RPA F1 & PA F1  \rule{0pt}{10pt}  \\
\midrule
\CheckmarkBold & \usym{2717}  & \usym{2717} & 5.55 & 31.68 &  11.30  &  81.61  \\
\usym{2717} & \CheckmarkBold & \usym{2717}  & 2.93 & 17.24 &  8.22  &  81.24  \\
\usym{2717} & \usym{2717} & \CheckmarkBold  & 12.47 & 46.45 &  34.04  &  86.12  \\
\midrule
\CheckmarkBold & \CheckmarkBold & \usym{2717}  & 27.19 & 58.65 & 26.09 & 84.27  \\
\CheckmarkBold & \usym{2717} & \CheckmarkBold  & 23.12 & 55.12 & 8.73 & 81.60  \\
\usym{2717} & \CheckmarkBold & \CheckmarkBold  & 17.44 & 46.18 & 3.92 & 74.52 \\
\hline
\CheckmarkBold & \CheckmarkBold & \CheckmarkBold  & \textbf{28.31} & \textbf{59.47} & \textbf{38.60} & \textbf{87.16} \\
\bottomrule
\end{tabular}

\label{ablation}
\end{table}

\begin{figure}[t] 
\centering
\subfigure[UCR]{\includegraphics[width=0.40\textwidth]{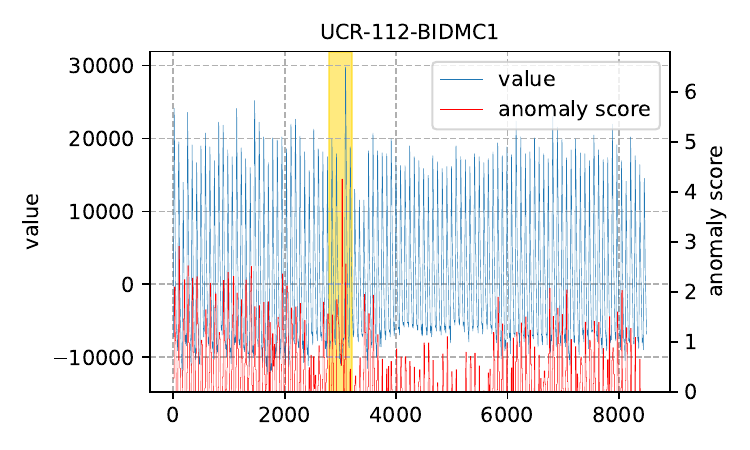}}
\subfigure[SWaT]{\includegraphics[width=0.40\textwidth]{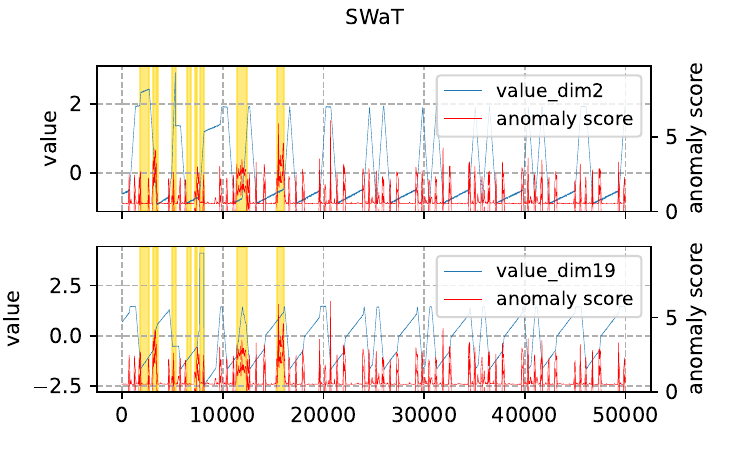}}
\caption{The visualization of \ourmethod{} AD results on UCR and SWaT datasets. The original data is present as the blue curves. The yellow areas are ground-truth anomalies, including point- and pattern-wise ones. The red curves represent the anomaly scores calculated by our method.}
\label{visualization}
\end{figure}

\subsection{Visualization}
To explain how our method works to detect diverse abnormal sequences and provide a more intuitive evaluation, the visualizations of the SensitiveHUE model with the \ourmethod{} learning objective on the univariate dataset UCR and the multivariate dataset SWaT are conducted, in Fig. \ref {visualization}. 
As shown in Fig. \ref{visualization}(a), we show the 115th BIDMC1 sequence in the UCR dataset, which contains a pattern-wise anomaly.
As observed in the figure, the anomaly scores exhibit significant increases around the ground-truth anomaly regions, closely aligning with the actual anomaly locations. This indicates that our \ourmethod{} method demonstrates strong detection performance and effectively captures local fluctuations.
Furthermore, in Fig. \ref{visualization}(b), which represents the SWaT dataset, the time series contains more complex pattern-wise anomalies, and we visualize two selected dimensions from the dataset.
Compared to the UCR dataset, although the anomaly regions in the SWaT dataset appear more dispersed, our method still produces clear fluctuations in anomaly scores near the anomalous regions, demonstrating \ourmethod{}’s sensitivity to pattern variations.
Although this method calculates a few other high-scoring segments, its overall detection performance is excellent in multivariate time series scenarios.
In conclusion, \ourmethod{} effectively addresses the limitations of other learning objectives on pattern-wise anomaly, which could enhance the performance of reconstruction-based methods.

\begin{figure}[t] 
\centering
\subfigure[$\lambda_{1},\lambda_2$(AIOps)]{\includegraphics[width=0.23\textwidth]{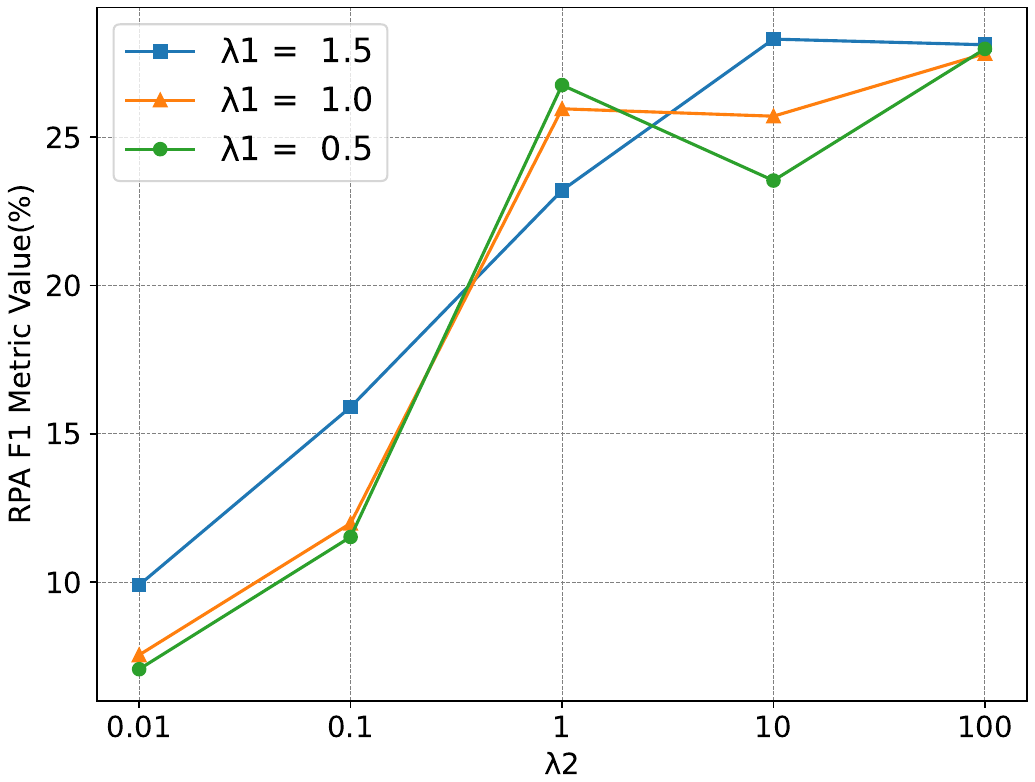}}
\subfigure[$\varepsilon$ (AIOps)]{\includegraphics[width=0.23\textwidth]{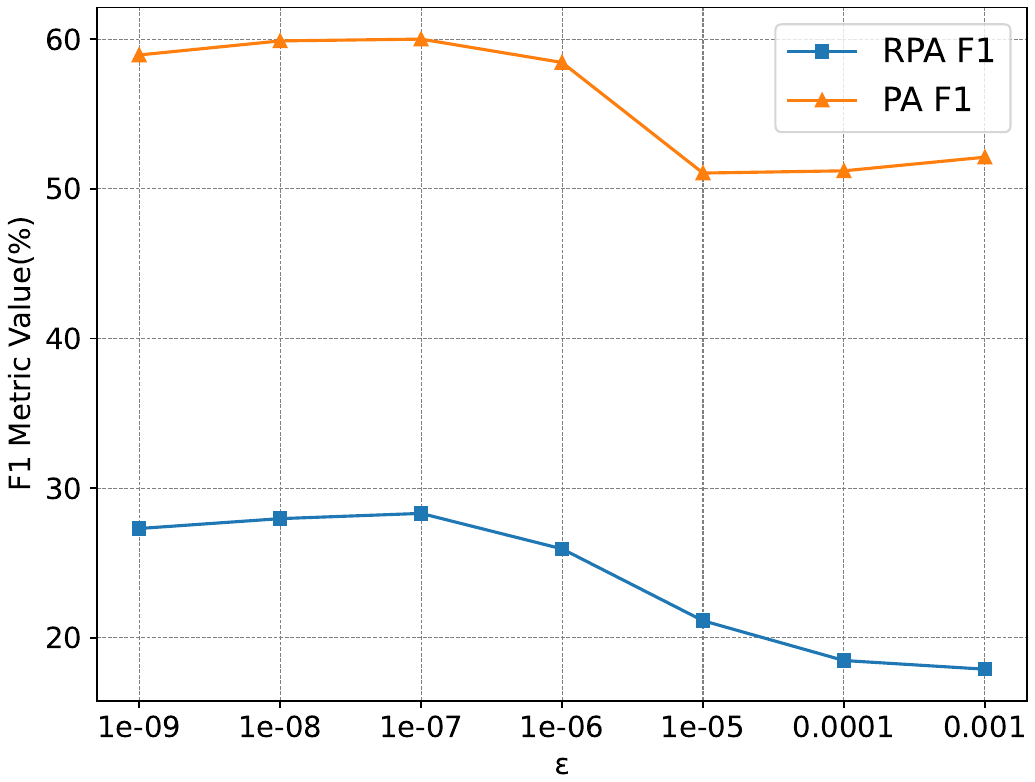}}
\caption{Sensitivity analysis on the AIOps dataset. }
\label{hyper}
\end{figure}

\subsection{Hyperparameters Analysis}
\label{sec:hyper}
As shown in Fig. \ref{hyper}, we conducted a sensitivity analysis to study hyperparameters' impact, including the weight parameter controlling $\lambda_1, \lambda_2$ in Eq. \ref{eq06}, and stability constant $\varepsilon$ in Eq. \ref{eq03}.
In Fig. \ref{hyper}(a), the y-axis represents the RPA F1 score, and the x-axis represents $\lambda_2 \in (0, 100] $.
The ablation study results demonstrate that the trend-seasonality-based variant achieves remarkable performance on the AIOps dataset. 
Accordingly, we conduct sensitivity experiments by setting three different values (0.5, 1.0, and 1.5) for the hyperparameter $\lambda_1$ to further investigate its impact on model performance.
As shown in Fig. \ref{hyper}(a), we observe that the ratio between parameters $\lambda_1$ and $\lambda_2$ should lie within a moderate range.
However, beyond a certain point, an excessively high or low ratio can reduce the \ourmethod{} performance.
we observe that the $\lambda_1 = 1.5 $ and $\lambda_2 = 10$ perform the best.
Therefore, each structural component of the \ourmethod{} optimization objective plays an indispensable role, and only an appropriately weighted combination of these components can lead to optimal performance.

Fig. \ref{hyper}(b) shows the results of varying the constant $\varepsilon$ in a range between 0 and 0.001.
The graph indicates that the \ourmethod{} learning objective performs best on the AIOps dataset in terms of RPA F1 score and PA F1 score when the stability constant $\varepsilon$ is set to 1e-7.
In the formulation Eq. \ref{eq03}, $\varepsilon$ is a small stability constant. 
When set too high, it may cause suspicious anomalies to be overlooked, thereby degrading overall detection performance—an observation that aligns with the variation trend of the RPA F1 score shown in Fig. \ref{hyper}(b). 
In contrast, the PA F1 score exhibits changes slightly under different values of $\varepsilon$, indicating that the RPA metric provides a fair and sensitive evaluation for time series anomaly detection compared with the PA metric.

\section{Conclusion}
We are among the first attempts to take into account the structure similarity to improve the optimization objective for time series anomaly detection models. This work details a practical solution \ourmethod{} that guides the models to capture structural features inherent in time series data. To do so, we extract structural characteristics of time series and break down structural elements into trend, seasonality, and shape before elaborating a dedicated loss function individually. As opposed to conventional point-by-point optimization objectives, \ourmethod{} promotes the alignment between the original data and the reconstructed data in terms of structural features, through a structure-aware optimization objective mechanism, keeping consistency in global fluctuation and local characteristics. By integrating structure similarity analysis, \ourmethod{} can improve the performance of TSAD methods by enhancing the sensitivity of both point- and pattern-wise anomalies.
We conduct extensive evaluations with three SOTA models on five public datasets to verify the effectiveness of \ourmethod{}.
The results show that \ourmethod{} is a flexible optimization objective that could improve the performance of reconstruction-based models.
As a plug-and-play strategy, the \ourmethod{} applies to any time series reconstruction method and enables the model to achieve higher results without architectural modifications.
We envision that \ourmethod{} could stimulate future research on the structural optimization objective of time series anomaly detection.


\bibliographystyle{IEEEbib}
\bibliography{ref}

\begin{thebibliography}{10}

\bibitem{schmidl2022anomaly}
Sebastian Schmidl, Phillip Wenig, and Thorsten Papenbrock,
\newblock ``Anomaly detection in time series: a comprehensive evaluation,''
\newblock {\em Proceedings of the VLDB Endowment}, vol. 15, no. 9, pp. 1779--1797, 2022.

\bibitem{lai2021revisiting}
Kwei-Herng Lai, Daochen Zha, Junjie Xu, Yue Zhao, Guanchu Wang, and Xia Hu,
\newblock ``Revisiting time series outlier detection: Definitions and benchmarks,''
\newblock in {\em Thirty-fifth conference on neural information processing systems datasets and benchmarks track (round 1)}, 2021.

\bibitem{zamanzadeh2024deep}
Zahra Zamanzadeh~Darban, Geoffrey~I Webb, Shirui Pan, Charu Aggarwal, and Mahsa Salehi,
\newblock ``Deep learning for time series anomaly detection: A survey,''
\newblock {\em ACM Computing Surveys}, vol. 57, no. 1, pp. 1--42, 2024.

\bibitem{pang2021deep}
Guansong Pang, Chunhua Shen, Longbing Cao, and Anton Van~Den Hengel,
\newblock ``Deep learning for anomaly detection: A review,''
\newblock {\em ACM computing surveys (CSUR)}, vol. 54, no. 2, pp. 1--38, 2021.

\bibitem{xu2022anomaly}
Jiehui Xu, Haixu Wu, Jianmin Wang, and Mingsheng Long,
\newblock ``Anomaly transformer: Time series anomaly detection with association discrepancy,''
\newblock in {\em International Conference on Learning Representations}, 2022.

\bibitem{mou2023deep}
Xudong Mou, Rui Wang, Tiejun Wang, Jie Sun, Bo~Li, Tianyu Wo, and Xudong Liu,
\newblock ``Deep autoencoding one-class time series anomaly detection,''
\newblock in {\em ICASSP 2023-2023 IEEE International Conference on Acoustics, Speech and Signal Processing (ICASSP)}. IEEE, 2023, pp. 1--5.

\bibitem{feng2024sensitivehue}
Yuye Feng, Wei Zhang, Yao Fu, Weihao Jiang, Jiang Zhu, and Wenqi Ren,
\newblock ``Sensitivehue: Multivariate time series anomaly detection by enhancing the sensitivity to normal patterns,''
\newblock in {\em Proceedings of the 30th ACM SIGKDD Conference on Knowledge Discovery and Data Mining}, 2024, pp. 782--793.

\bibitem{malhotra2016lstm}
Pankaj Malhotra, Anusha Ramakrishnan, Gaurangi Anand, Lovekesh Vig, Puneet Agarwal, and Gautam Shroff,
\newblock ``Lstm-based encoder-decoder for multi-sensor anomaly detection,''
\newblock {\em arXiv preprint arXiv:1607.00148}, 2016.

\bibitem{wang2023deep}
Rui Wang, Chongwei Liu, Xudong Mou, Kai Gao, Xiaohui Guo, Pin Liu, Tianyu Wo, and Xudong Liu,
\newblock ``Deep contrastive one-class time series anomaly detection,''
\newblock in {\em Proceedings of the 2023 SIAM International Conference on Data Mining (SDM)}. SIAM, 2023, pp. 694--702.

\bibitem{wang2024deep}
Jun Wang, Wenjie Du, Yiyuan Yang, Linglong Qian, Wei Cao, Keli Zhang, Wenjia Wang, Yuxuan Liang, and Qingsong Wen,
\newblock ``Deep learning for multivariate time series imputation: A survey,''
\newblock {\em arXiv preprint arXiv:2402.04059}, 2024.

\bibitem{mou2025roca}
Xudong Mou, Rui Wang, Bo~Li, Tianyu Wo, Jie Sun, Hui Wang, and Xudong Liu,
\newblock ``Roca: Robust contrastive one-class time series anomaly detection with contaminated data,''
\newblock {\em arXiv preprint arXiv:2503.18385}, 2025.

\bibitem{tuli2022tranad}
Shreshth Tuli, Giuliano Casale, and Nicholas~R Jennings,
\newblock ``Tranad: Deep transformer networks for anomaly detection in multivariate time series data,''
\newblock {\em arXiv preprint arXiv:2201.07284}, 2022.

\bibitem{wang2023drift}
Chengsen Wang, Zirui Zhuang, Qi~Qi, Jingyu Wang, Xingyu Wang, Haifeng Sun, and Jianxin Liao,
\newblock ``Drift doesn't matter: dynamic decomposition with diffusion reconstruction for unstable multivariate time series anomaly detection,''
\newblock {\em Advances in Neural Information Processing Systems}, vol. 36, pp. 10758--10774, 2023.

\bibitem{gao2024units}
Shanghua Gao, Teddy Koker, Owen Queen, Thomas Hartvigsen, Theodoros Tsiligkaridis, and Marinka Zitnik,
\newblock ``Units: Building a unified time series model,''
\newblock {\em arXiv e-prints}, pp. arXiv--2403, 2024.

\bibitem{zhu2024llms}
Jiaqi Zhu, Shaofeng Cai, Fang Deng, Beng~Chin Ooi, and Junran Wu,
\newblock ``Do llms understand visual anomalies? uncovering llm's capabilities in zero-shot anomaly detection,''
\newblock in {\em Proceedings of the 32nd ACM International Conference on Multimedia}, 2024, pp. 48--57.

\bibitem{zhou2023one}
Tian Zhou, Peisong Niu, Liang Sun, Rong Jin, et~al.,
\newblock ``One fits all: Power general time series analysis by pretrained lm,''
\newblock {\em Advances in neural information processing systems}, vol. 36, pp. 43322--43355, 2023.

\bibitem{wang2024cutaddpaste}
Rui Wang, Xudong Mou, Renyu Yang, Kai Gao, Pin Liu, Chongwei Liu, Tianyu Wo, and Xudong Liu,
\newblock ``Cutaddpaste: Time series anomaly detection by exploiting abnormal knowledge,''
\newblock in {\em Proceedings of the 30th ACM SIGKDD Conference on Knowledge Discovery and Data Mining}, 2024, pp. 3176--3187.

\bibitem{xu2015learning}
Dan Xu, Elisa Ricci, Yan Yan, Jingkuan Song, and Nicu Sebe,
\newblock ``Learning deep representations of appearance and motion for anomalous event detection,''
\newblock {\em arXiv preprint arXiv:1510.01553}, 2015.

\bibitem{lu2017unsupervised}
Weining Lu, Yu~Cheng, Cao Xiao, Shiyu Chang, Shuai Huang, Bin Liang, and Thomas Huang,
\newblock ``Unsupervised sequential outlier detection with deep architectures,''
\newblock {\em IEEE transactions on image processing}, vol. 26, no. 9, pp. 4321--4330, 2017.

\bibitem{su2019robust}
Ya~Su, Youjian Zhao, Chenhao Niu, Rong Liu, Wei Sun, and Dan Pei,
\newblock ``Robust anomaly detection for multivariate time series through stochastic recurrent neural network,''
\newblock in {\em Proceedings of the 25th ACM SIGKDD international conference on knowledge discovery \& data mining}, 2019, pp. 2828--2837.

\bibitem{wang2024fredf}
Hao Wang, Licheng Pan, Zhichao Chen, Degui Yang, Sen Zhang, Yifei Yang, Xinggao Liu, Haoxuan Li, and Dacheng Tao,
\newblock ``Fredf: Learning to forecast in frequency domain,''
\newblock {\em arXiv preprint arXiv:2402.02399}, 2024.

\bibitem{kudrat2025patch}
Dilfira Kudrat, Zongxia Xie, Yanru Sun, Tianyu Jia, and Qinghua Hu,
\newblock ``Patch-wise structural loss for time series forecasting,''
\newblock {\em arXiv preprint arXiv:2503.00877}, 2025.

\bibitem{lai2023nominality}
Chih-Yu~Andrew Lai, Fan-Keng Sun, Zhengqi Gao, Jeffrey~H Lang, and Duane Boning,
\newblock ``Nominality score conditioned time series anomaly detection by point/sequential reconstruction,''
\newblock {\em Advances in Neural Information Processing Systems}, vol. 36, pp. 76637--76655, 2023.

\bibitem{AIOpsChallenge}
``Aiops challenge. the 1st match for aiops.,'' \url{https://github.com/NetManAIOps/KPI-Anomaly-Detection}, 2018,
\newblock Accessed: 2024-12-04.

\bibitem{wu2021current}
Renjie Wu and Eamonn~J Keogh,
\newblock ``Current time series anomaly detection benchmarks are flawed and are creating the illusion of progress,''
\newblock {\em IEEE transactions on knowledge and data engineering}, vol. 35, no. 3, pp. 2421--2429, 2021.

\bibitem{mathur2016swat}
Aditya~P Mathur and Nils~Ole Tippenhauer,
\newblock ``Swat: A water treatment testbed for research and training on ics security,''
\newblock in {\em 2016 international workshop on cyber-physical systems for smart water networks (CySWater)}. IEEE, 2016, pp. 31--36.

\bibitem{ahmed2017wadi}
Chuadhry~Mujeeb Ahmed, Venkata~Reddy Palleti, and Aditya~P Mathur,
\newblock ``Wadi: a water distribution testbed for research in the design of secure cyber physical systems,''
\newblock in {\em Proceedings of the 3rd international workshop on cyber-physical systems for smart water networks}, 2017, pp. 25--28.

\bibitem{kotowski2024european}
Krzysztof Kotowski, Christoph Haskamp, Jacek Andrzejewski, Bogdan Ruszczak, Jakub Nalepa, Daniel Lakey, Peter Collins, Aybike Kolmas, Mauro Bartesaghi, Jose Martinez-Heras, et~al.,
\newblock ``European space agency benchmark for anomaly detection in satellite telemetry,''
\newblock {\em arXiv preprint arXiv:2406.17826}, 2024.

\bibitem{sakoe1978dynamic}
Hiroaki Sakoe and Seibi Chiba,
\newblock ``Dynamic programming algorithm optimization for spoken word recognition,''
\newblock {\em IEEE transactions on acoustics, speech, and signal processing}, vol. 26, no. 1, pp. 43--49, 1978.

\bibitem{lee2021soft_dtw_loss}
Keon Lee,
\newblock ``Soft-dtw-loss,'' \url{https://github.com/keonlee9420/Soft-DTW-Loss}, 2021.

\bibitem{bhatnagar2021merlion}
Aadyot Bhatnagar, Paul Kassianik, Chenghao Liu, Tian Lan, Wenzhuo Yang, Rowan Cassius, Doyen Sahoo, Devansh Arpit, Sri Subramanian, Gerald Woo, et~al.,
\newblock ``Merlion: A machine learning library for time series. 2021,''
\newblock {\em URL https://arxiv. org/abs/2109.09265}, 2021.

\bibitem{kim2022towards}
Siwon Kim, Kukjin Choi, Hyun-Soo Choi, Byunghan Lee, and Sungroh Yoon,
\newblock ``Towards a rigorous evaluation of time-series anomaly detection,''
\newblock in {\em Proceedings of the AAAI Conference on Artificial Intelligence}, 2022, vol.~36, pp. 7194--7201.

\end{thebibliography}

\end{document}